\documentclass{IEEEcsmag}

\usepackage[colorlinks,urlcolor=blue,linkcolor=blue,citecolor=blue]{hyperref}

\usepackage{upmath}

\jvol{XX}
\jnum{XX}
\paper{8}

\usepackage{algorithm}
\usepackage{gensymb}
\usepackage{hyperref}
\usepackage{graphicx} 
\usepackage{subcaption}
\usepackage{algorithmic}
\renewcommand{\algorithmicrequire}{\textbf{Input:}}
\renewcommand{\algorithmicensure}{\textbf{Output:}}
\usepackage{verbatim}

\begin{document}

\title{Optimizing Drone Delivery in Smart Cities}

\author{\IEEEauthorblockN{Babar Shahzaad, Balsam Alkouz, Jermaine Janszen, Athman Bouguettaya
}

\IEEEauthorblockA{School of  Computer Science,
The University of Sydney, Australia
}
}

\markboth{}{Optimizing Drone Delivery in Smart Cities}

\begin{abstract}
We propose a novel context-aware drone delivery framework for {\em optimizing} package delivery through skyway networks in smart cities. We reformulate the problem of finding an optimal drone service delivery pathway as a more {\em congruent} and {\em elegant} drone delivery service {\em composition} problem. In this respect, we propose a novel line-of-sight {\em heuristic-based} context-aware composition algorithm that selects and composes near-optimal drone delivery services. We conducted an extensive experiment using a real dataset to show the robustness of our proposed approach.

\end{abstract}

\maketitle

\chapterinitial{Unmanned Aerial Vehicles (UAVs)} are gaining increasing attention as means for service provisioning in smart cities \cite{7807172}. A drone is a popular type of UAV that offers potential benefits in smart city applications \cite{9707164}. Drones are increasingly becoming pervasive in their use, including surveillance, agriculture, and delivery of goods \cite{chmaj2015distributed}. During the COVID-19 pandemic, drones have been widely used for monitoring social distancing, aerial spraying, and delivery of goods. Several countries have used drones for safe and contactless deliveries during pandemic lockdowns \cite{KUMAR20211}. Moreover, companies such as Amazon and Google have massively increased their investment in drones for delivery services \cite{Aurambout2019}. Drone delivery services are typically targeted at consumers and suppliers of goods and services (e.g., retailers, pharmaceutical and medical suppliers, and transport services).

The \emph{service paradigm} is congruent with the concept of drone service delivery \cite{9380171}. It provides an elegant mechanism to define and model the drone's functional and non-functional aspects as a drone delivery {\em service}. In this respect, the functional property depicts the {\em delivery of a package} from one node to another while traversing a {\em skyway} network \cite{shahzaad2021top}. A drone's maximum payload weight, battery capacity, and flying range represent the non-functional (aka \textit{Quality of Service} (QoS)) aspects. A {\em skyway network} is defined as a set of skyway \textit{segments} that directly connect two nodes representing take-off and landing stations \cite{10.1145/3460418.3479289}. Take-off and landing stations (aka network nodes) are typically from and to building rooftops. Each network node acts as a {\em delivery target} and a {\em recharging station}. In this regard, an {\em atomic} drone delivery service is characterized by the transportation of a package using a drone along a skyway segment operating under a set of constraints.

\begin{figure}

    \centering
    \includegraphics[width=\linewidth]{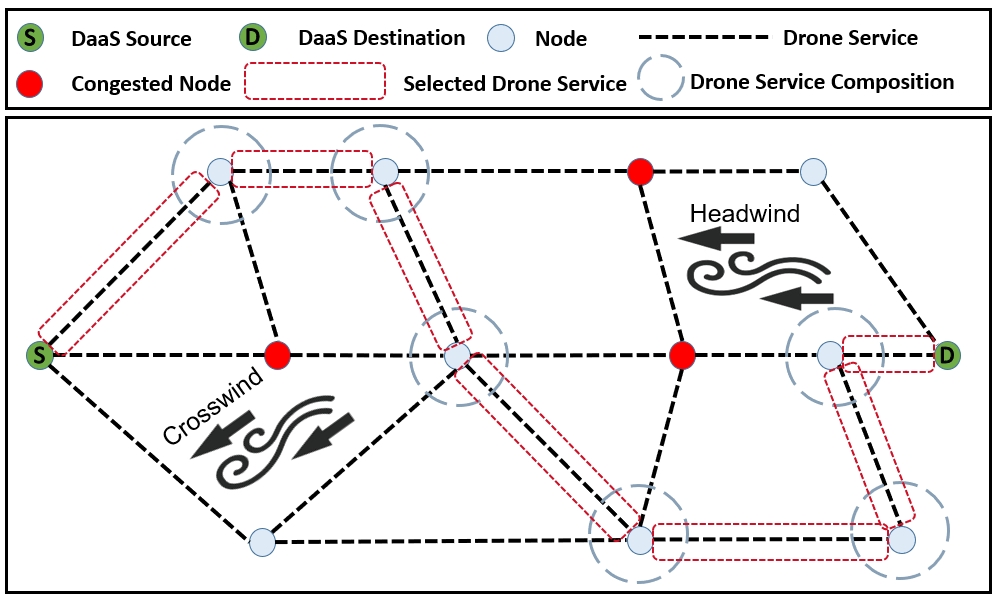}
    \caption{Drone Service Composition Scenario}
    \label{fig1}

\end{figure}

Delivery drones are typically constrained by their battery capacity and flying range \cite{5}. For example, the DJI M200 V2 drone can only fly as far as 32 km when fully charged. Several solutions have been proposed to address these constraints, such as battery swapping \cite{cokyasar2021designing}, solar-powered batteries, and the use of additional batteries \cite{galkin2019uavs}. However, these solutions either require a precise landing of drones, availability of spare batteries specific to drones, or are highly dependent on the weather conditions. The recharging of drones at intermediate stations is an applicable solution in the context of utilizing building rooftops as recharging stations \cite{huang2021deployment}. In this regard, a drone may need to be recharged multiple times to cover long-distance areas. In service terms, this would translate into a service {\em composition} to deliver goods. An \textit{optimal drone delivery service composition} is the process of selecting and composing the {\em temporally optimal} drone delivery services from a given source location to a delivery destination \cite{9707164}. In this context, temporally optimal refers to leading towards the destination faster. {\bf Figure \ref{fig1}} provides a depiction of a drone service composition scenario.

Drone service composition broadly involves two types of constraints: (1) internal/inherent constraints and (2) external/environmental constraints \cite{jermaine2021demo}. The internal/inherent constraints include limited payload weight, flying range, and limited battery capacity of a drone. The external/environmental constraints include recharging pad availability, congestion at stations, and wind conditions. Existing approaches mainly focus on the last-mile delivery services \cite{8798337} or delivery in deterministic environments \cite{8818436}. In these approaches, assumptions are simplified to ignore the impact of the constraints mentioned above on drone delivery plans. In contrast, we make no such simplifying assumptions in our work. We consider the internal and external constraints of drone-based provisioning of delivery.

We propose a context-aware drone delivery framework for effectively provisioning delivery services in smart cities. In this paper, context-awareness refers to the capability of leveraging internal/inherent and external/environmental constraints information to tailor the near-optimal drone service composition. We develop a Line-of-Sight (LOS) heuristic composition algorithm that selects and composes the best drone delivery services. A LOS heuristic is based on a straight line from the drone's current location to the target location to determine the next optimal node to visit. In this respect, the best drone service is a service that guarantees package delivery in a minimum time from its start to end location. Using a real dataset, we conduct experiments to demonstrate the effectiveness of our LOS heuristic approach.

\section{Related Work} \label{relatedwork}

Existing drone delivery planning and scheduling approaches can be divided into point-to-point and multi-point delivery approaches \cite{9707164}. In point-to-point delivery approaches, the deliveries are made in a limited geographical area. In such cases, a drone takes off from the warehouse, delivers the package directly to the customer, and returns to the warehouse. A job assignment problem is studied to dimension and control a fleet of drones in a drone delivery system \cite{grippa2019drone}. Two policies are proposed using queuing theory to solve the job assignment problem. The first policy
uses the customer's location to the jobs. The second policy uses the arrival time of the customer's request to select the jobs. Simulation experiments show that the second policy is more effective in optimizing the delivery time for low loads and performs well for high loads.

A drone service framework is proposed to provide delivery services \cite{9380171}.
Scheduling, route planning, and composition are the fundamental components of the proposed framework. The scheduling generates itineraries for drones in a network. A route-planning algorithm is proposed focusing on the selection of the optimal route. The drone services are composed at each station using a drone service composition algorithm. \textit{The proposed framework does not consider the LOS drone flying regulations and drones' recharging requirements.}

Our previous studies are the first to model the multi-point drone deliveries using the service paradigm \cite{8818436,shahzaad2021top}. A Drone-as-a-Service (DaaS) framework is presented for package delivery using drones \cite{8818436}. This paper was the first to model a drone's capabilities as a DaaS. This study aims to select and compose the right set of DaaS for faster package delivery. A heuristic-based algorithm is presented for DaaS selection and composition. \textit{The proposed composition approach considers the environment to be static and ignores the recharging constraints, which is not realistic in practical scenarios.}

A dynamic top-k service composition approach is proposed for drone delivery services \cite{shahzaad2021top}. Two drone service models are presented considering no-congestion and with congestion conditions. The no-congestion drone service model ignores the congestion conditions and computes the initial top-k compositions. This initial plan is updated after incorporating congestion conditions using the congestion model.
The impact of wind conditions is not addressed in the proposed approach. To the extent of our knowledge, none of the existing approaches consider the effects of the relative wind (i.e., headwind, tailwind, and crosswind) on the drone's energy consumption. This paper is the first attempt to propose a context-aware drone delivery framework that considers the congestion conditions at recharging stations and the effect of relative wind on the drone's energy consumption. 

\section{Context-Aware Drone Delivery Framework}

\subsection{Multi-Drone Skyway Network}

This section describes our multi-drone skyway network, where multiple drones are assumed to operate to deliver packages to respective delivery targets. We focus solely on using drones to perform deliveries through a skyway network in a given geographical area. A skyway network enables the real-world deployment of drone delivery systems in integrated airspace. We construct a multi-drone skyway network by linking predefined skyway segments between any two nodes within the line of sight. Each node has a landing pad on a residential/commercial rooftop. Our proposed skyway network uses a city's existing infrastructure where wireless recharging pads can easily and cheaply be fitted on building rooftops.

We formally define a drone delivery service and request for drone delivery as follows.

\textbf{Definition 1:} We define a Drone Delivery Service (DDS) as a tuple $<DDS\_id,$ $DDS_{f}, DDS_q>$, where
\begin{itemize}
    \item[$\bullet$] $DDS.id$ represents a unique ID,
    \item[$\bullet$] $DDS_{f}$ is the delivery function of a drone $D$ to transport a package over a skyway segment,
    \item[$\bullet$] $DDS_q$ is a set of QoS attributes of a drone delivery service.
\end{itemize}

\textbf{Definition 2:} A drone delivery service request is a tuple $<\zeta, \xi, rt_{s}, w>$, where
\begin{itemize}
    \item[$\bullet$] $\zeta$ represents the source (i.e., warehouse),
    \item[$\bullet$] $\xi$ represents the destination (i.e., customer location),
    \item[$\bullet$] $rt_s$ represents the start time of the request,
    \item[$\bullet$] $w$ represents the weight of the package to be delivered.
\end{itemize}

\subsection{Problem Formulation}

Given a drone-based delivery service request from a consumer, we formulate the drone delivery problem as the composition of the best $DDS$ to form a skyway path from source $\zeta$ to destination $\xi$. The aforementioned internal and external constraints pose significant challenges to the composition of drone delivery services. First, the selection of the right drone service to serve the delivery request. This selection process involves the consideration of both internal and external constraints for efficient service delivery. Second, the selection of subsequent drone services if a single drone service does not satisfy the delivery request due to battery limitations or line-of-sight drone flying regulations. Third, avoiding congestion conditions at recharging stations and selecting a wind-favored skyway path for faster service delivery. The objective of this problem formulation is to select and compose the drone services with the overall shortest delivery time from source to destination while considering all the aforementioned constraints and challenges.

\subsection{Line-of-Sight Heuristic Composition} 
We propose a LOS heuristic composition algorithm using an adapted A* algorithm to compose near-optimal services from source to destination. In Algorithm \ref{heuristic_alg}, a consumer-invoked service request includes the source node, destination node, and package weight. At the source node, the algorithm computes the drone's probability $P(A)$ to arrive at the destination straightaway without stopping at intermediate stations. $P(A)$ is computed using the path with the shortest distance to the destination. If the drone can fly this path and reach the destination directly before battery depletion, given the wind and payload constraint, then the value of $P(A)$ is 1 and the drone goes directly to the destination using this path. We assume to have a global wind direction $\theta$ and speed affecting all the skyway network segments. As shown in Equation \ref{rel_wind_eq}, we compute the relative wind impact on the drone ($\alpha$) by subtracting the drone's orientation ($\beta$) from the global wind direction ($\theta$).
We convert any $\alpha$ value to the ranges (0\degree \ to 180\degree) and (0\degree \ to -180\degree) (Lines \ref{180_range} -\ref{180_range_end}). Since the drone travels in a line-of-sight, we compute the drone's orientation using the positions of the two nodes connecting the segment (Equation \ref{orientation_eq}). Wind impact on a drone is typically modeled in three ways: headwind, tailwind, and crosswind \cite{chu2021simulation}. The Real Dataset, which is described in the following section, captures the effect of these types on the drone's energy consumption. However, because drones fly in a skyway network, the orientation of the drone may differ from the energy-favorable wind direction. Thus, our formula captures the effect of relative wind direction, which includes the orientation of the drone $\beta$ and the global wind direction $\theta$. We compute the arriving probability using the shortest path from source to destination in terms of distance only. The probability considers the payload and wind constraints at each segment. If the drone cannot reach the destination straightaway, the drone goes to the best neighboring recharging node (Lines \ref{bestneighbor}-\ref{endbestneighbor}).

\begin{equation} 
    \label{rel_wind_eq} 
    \alpha=\theta-\beta 
\end{equation} 

\begin{equation} 
    \label{orientation_eq} 
    \beta= atan2 (y2-y1, x2-x1) 
\end{equation} 

\begin{algorithm} [t]
 \caption{LOS Heuristic Composition Algorithm} 
 \label{heuristic_alg} 
 \small 
 \begin{algorithmic}[1] 
 \renewcommand{\algorithmicrequire}{\textbf{Input:}} 
 \renewcommand{\algorithmicensure}{\textbf{Output:}} 
 \REQUIRE $D$, $R$, $\theta$ 
 \ENSURE  $dt$ 
 \STATE $dt$ = 0 
    \WHILE{$D$ is not at destination} 
        \STATE path to destination= \textbf{Dijkstra}(current, destination) 
        \STATE distance to destination=0 
        \FOR{Every segment $s$ in the path to destination} 
            \STATE distance to destination+=$distance_{s_i}$ 
            \STATE $\beta_{s_i} = atan2 (s_{y_{i+1}}-s_{y_{i}},  \ s_{x_{i+1}}-s_{x_{i}})$ 
            \STATE $\alpha_{s_i}=\theta-\beta_{s_i}$ 
            \IF{$\alpha_{s_i} > 180$} \label{180_range} 
            \STATE $\alpha_{s_i} -= 360$ 
            \ELSIF{$\alpha_{s_i}<-180$} 
            \STATE $\alpha_{s_i} += 360$ 
            \ENDIF\label{180_range_end} 
        \ENDFOR 
        \STATE \textbf{compute} energy consumption for $D$ based on $R$ package weight, distance to destination, and $\alpha$ 
        \IF{$P(A)= 1$} 
        \STATE $D$ can reach the destination without intermediate nodes 
        \STATE $D$ travels to the destination 
        \STATE $dt$+=travel time 
        \ELSE 
        \STATE \textbf{find} nearest neighbor nodes \label{bestneighbor}
        \STATE \textbf{select} best neighboring node min($tt$+$rt$+$wt$+$ett$)) 
        \STATE $S$ travels to the neighboring node 
        \STATE $dt$ += $tt$ + $rt$ + $wt$ \label{endbestneighbor}
        \ENDIF 
    \ENDWHILE 
 \RETURN $dt$ 
 \end{algorithmic} 
 \end{algorithm} 

The best node is a node with the least travel and transit times. Transit time consists of the drone's recharging time $rt$ and waiting time $wt$ if a recharging pad is occupied.
Each segment's energy consumption rate and travel time are extracted from the dataset described in the Real Dataset section based on the segment's length, package weight, and wind conditions. When selecting the best neighbor, the euclidean distance and euclidean travel time $ett$ between each neighbor and the final destination is considered as a heuristic that adds to the travel time and transit time factors. Including the euclidean travel time factor ensures that a neighbor in the direction of the destination is selected and that the algorithm \textit{converges}. When the drone travels to the best neighbor, the delivery time is updated with the travel time of the segment and the transit time. The algorithm again checks if the drone can reach the destination directly. The process continues until the drone successfully reaches its destination. Because the proposed algorithm is a modified version of A*, the worst-case complexity is $O(|E|)=O(b^d)$. However, the heuristic function has a significant impact on the practical performance of the search because it allows the algorithm to prune away many of the $b^d$ nodes that an uninformed search would expand. Furthermore, because our algorithm can skip nodes when applicable, the complexity is typically reduced.

\section{Real Dataset} \label{dataset_collection} 
\begin{figure}
    \begin{subfigure}[b]{1\columnwidth}
        \includegraphics[width=\textwidth]{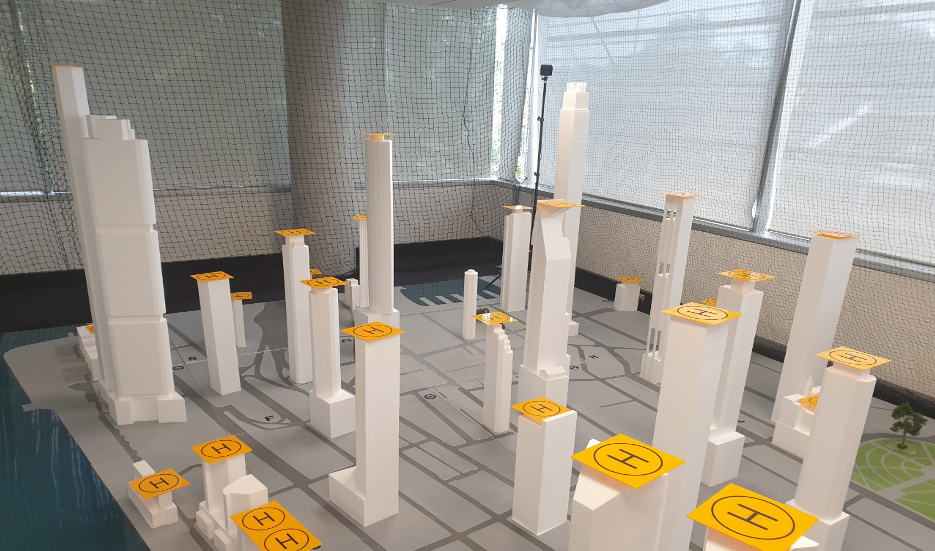} 
        \caption{Indoor Drone Testbed of Sydney’s CBD}
        \label{testbed-diagram}
    \end{subfigure}
    \hfill
    \begin{subfigure}[b]{1\columnwidth}
        \includegraphics[width=\textwidth]{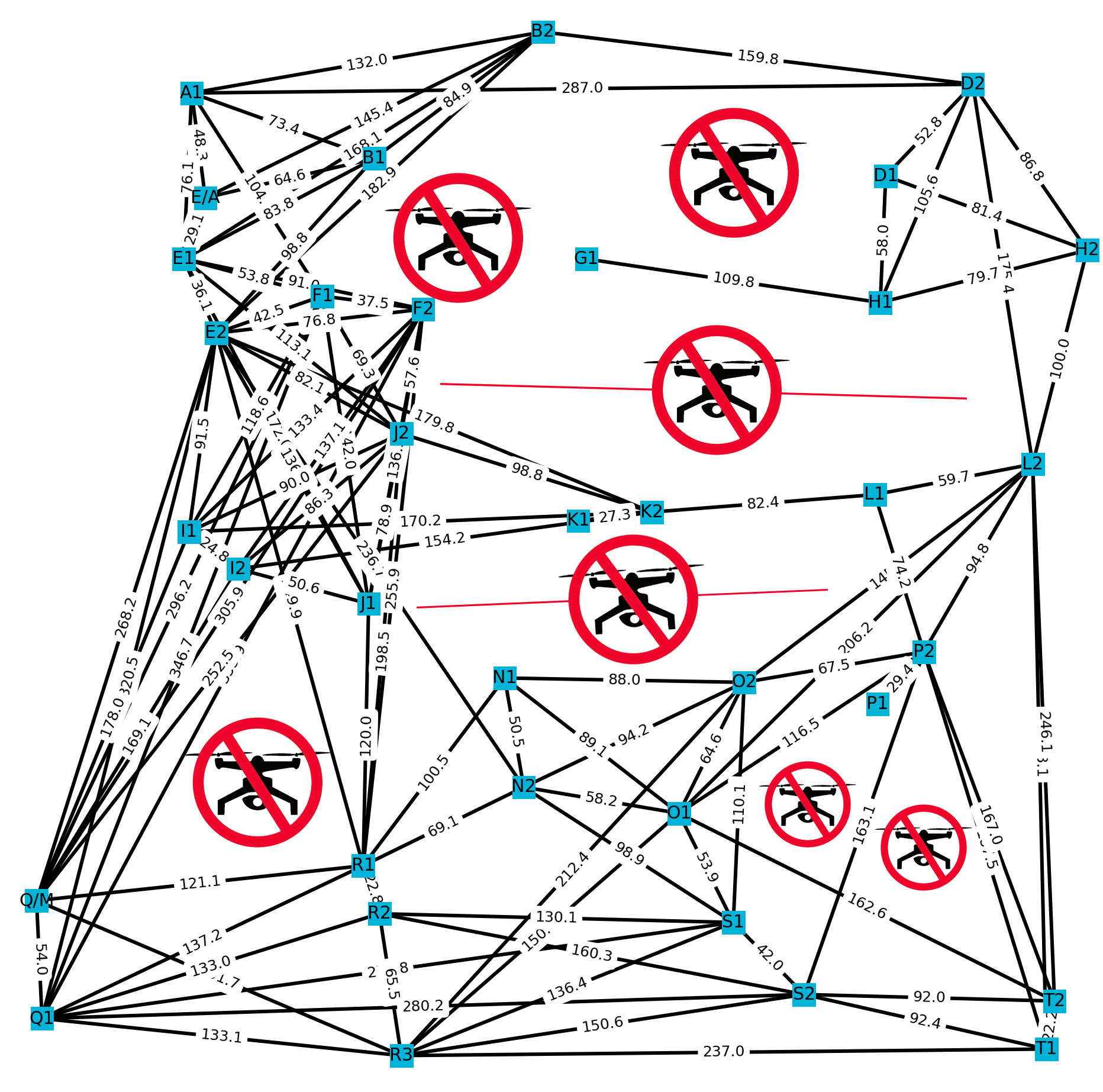} 
        \caption{Skyway Network with No-Flight Zones}
        \label{network-diagram} 
    \end{subfigure}

    \caption{3D Model of Sydney's CBD and Skyway Network}
    \label{fig:paths}
\end{figure}

We use a real dataset collected on various drone parameters \cite{jermaine2021demo}. An indoor drone testbed is set up using a 3D printed model of Sydney's CBD to construct a skyway network ({\bf Figure \ref{testbed-diagram}}). A drone would typically traverse building rooftops equipped with recharging pads to serve the delivery request. We use HTC Vive base stations to assist the drone in locating its precise position during the flight. The base stations are fitted at diagonal corners of the lab. We use Fanco Premium Pedestal fan to generate wind speeds in different directions. In real-world scenarios, various wind effects affect urban environments, such as venturi, downdraught, downwind eddy, counter-current effect, etc \cite{kang2020computational}. However, at any given time, the drone is primarily exposed to wind from one direction, i.e., headwind, tailwind, or crosswind. Therefore, we investigate the impact of wind direction on a drone scale rather than an urban scale, i.e., the major wind hitting the drone at any point in time. Furthermore, a city would typically include one weather base station that reports global wind speed and direction for practical reasons. It is not always possible to obtain microclimate data at the urban or building level. As a result, our proposed solution models wind impact on a drone by comparing the global wind direction to the drone's orientation. We construct a real miniature skyway network using the building rooftops of Sydney’s CBD ({\bf Figure \ref{network-diagram}}). The network consists of 36 nodes and seven no-flight zones following aviation regulations for safety and privacy purposes. The collected dataset records the impact of wind conditions and different payloads on the drone's energy consumption over a set of trajectory patterns. The trajectory patterns include \textit{hovering, linear, rectangular,} and \textit{triangular} flight paths. This dataset contains various attributes of a drone during each flight, including XY positions, altitude, energy consumption, and wind speed and direction. The data are stored in the form of CSV files\footnote{shorturl.at/cGLSV}. 

\begin{figure*}
    \begin{subfigure}[b]{1\columnwidth}
        \includegraphics[width=\textwidth]{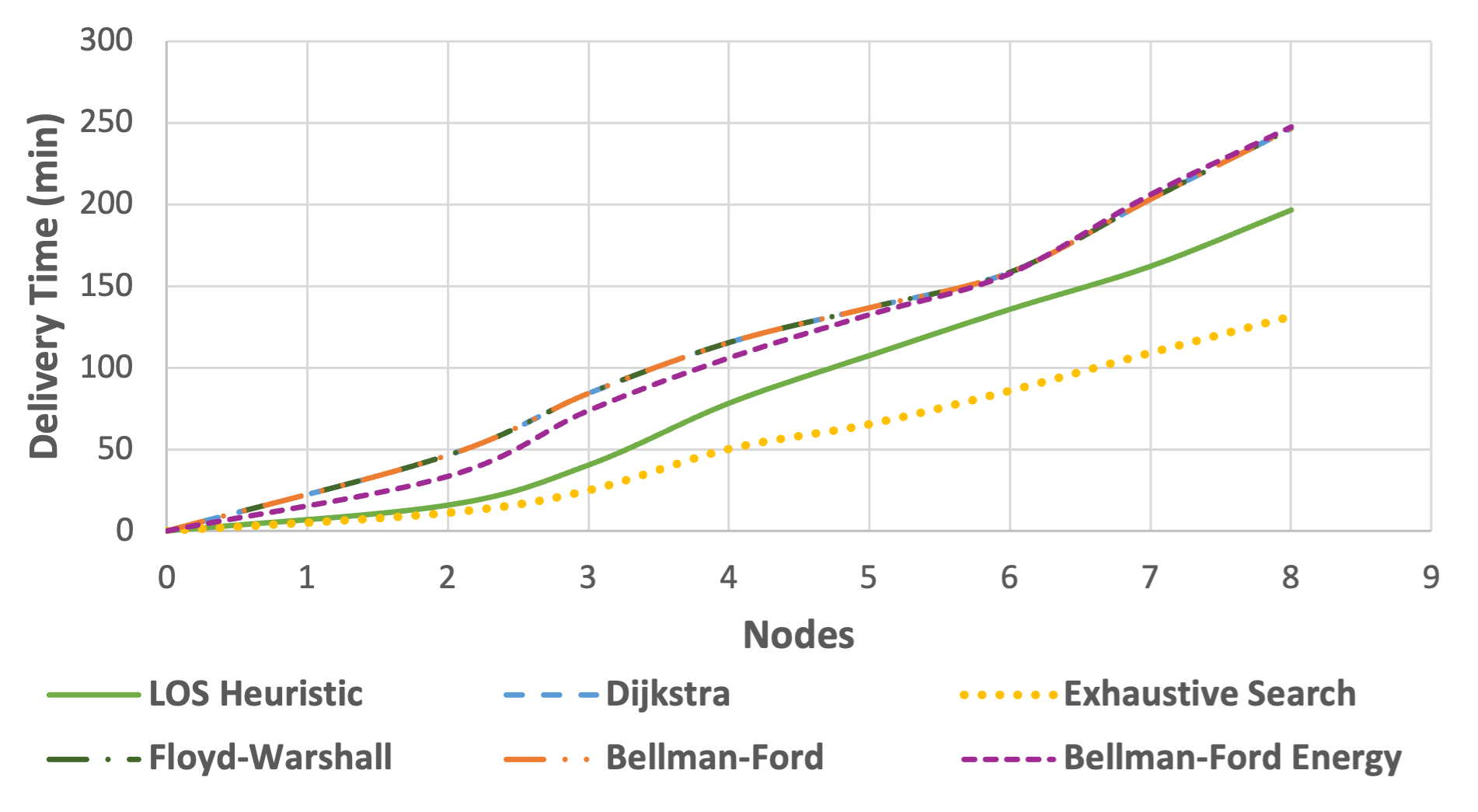}
        \caption{Average Delivery Time}
        \label{delivery_time}
    \end{subfigure}
    \hfill
    \begin{subfigure}[b]{1\columnwidth}
        \includegraphics[width=\textwidth]{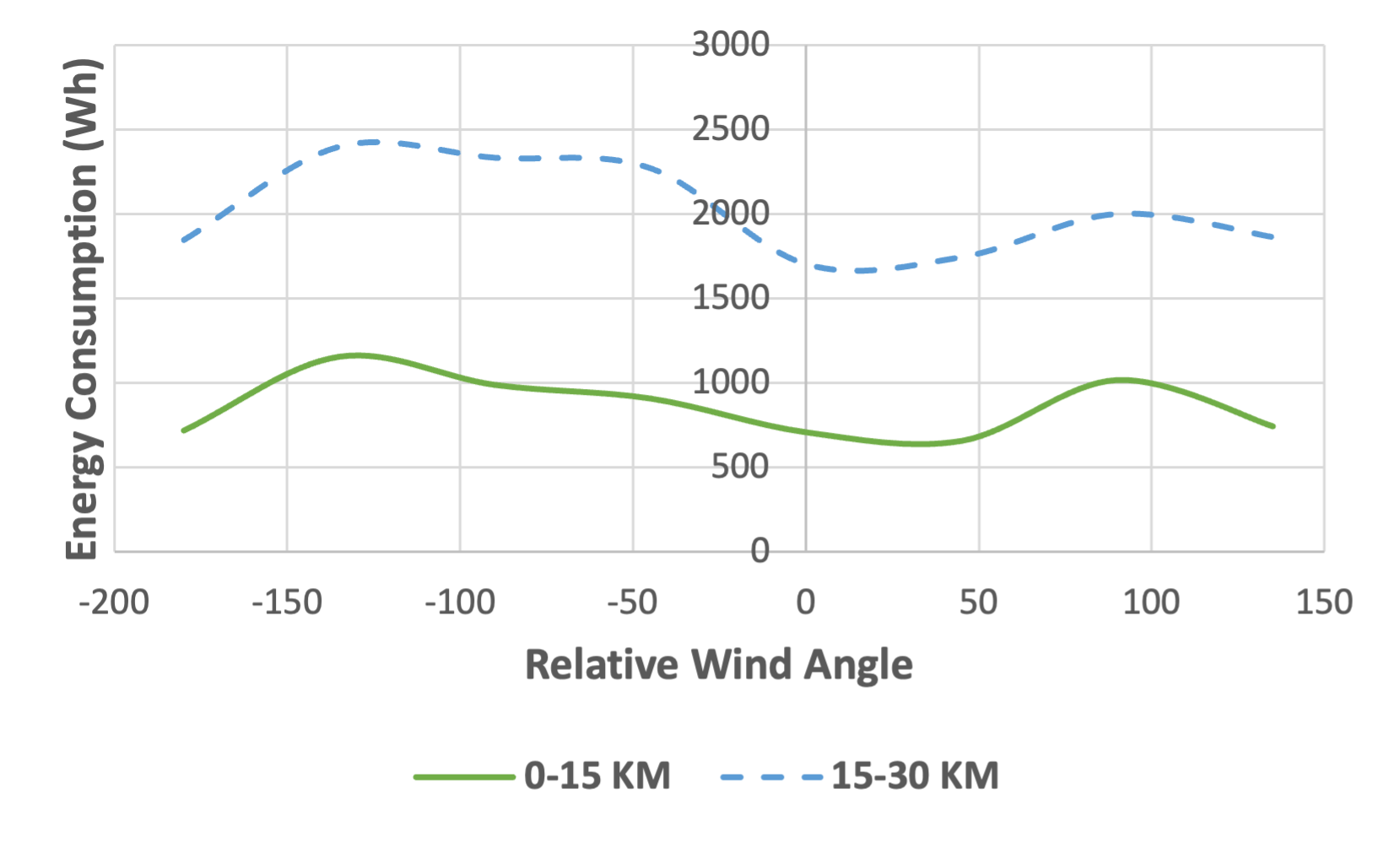}
        \caption{Effect of Relative Wind Directions on the Energy Consumption}
        \label{wind-energy}
    \end{subfigure}
    \hfill
    \begin{subfigure}[b]{1\columnwidth}
        \includegraphics[width=\textwidth]{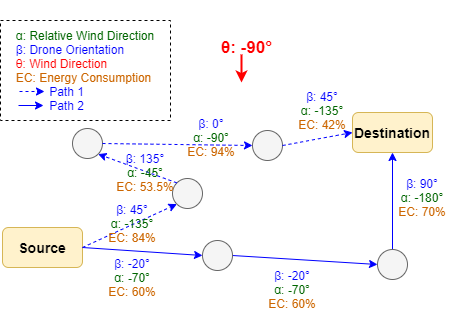}
        \caption{Path Composition with Relative Wind Direction}
        \label{wind-composition}
    \end{subfigure}
    \hfill
    \begin{subfigure}[b]{1\columnwidth}
        \includegraphics[width=\textwidth]{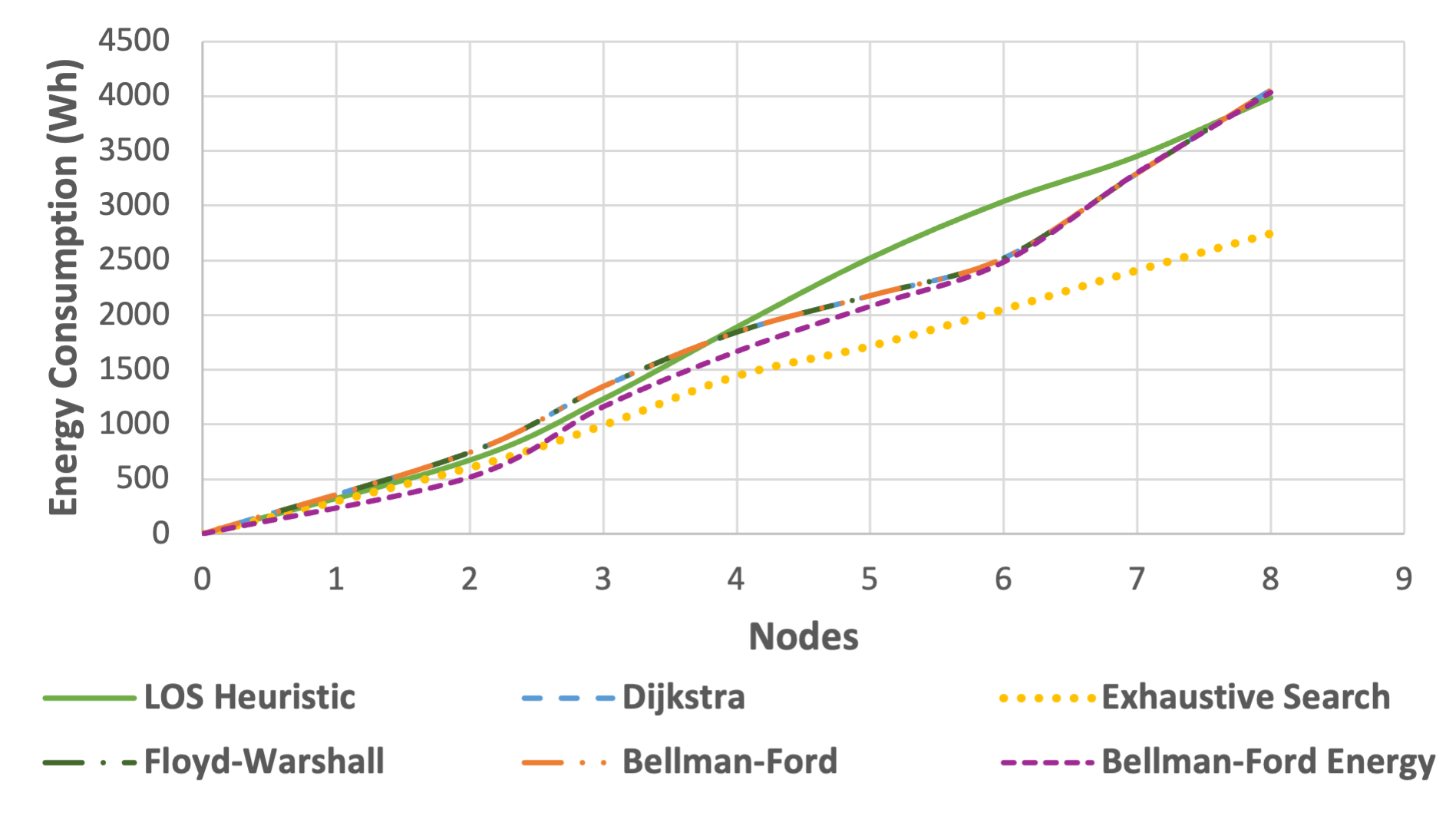}
        \caption{Average Energy Consumption}
        \label{energy_consumptuin}
    \end{subfigure}
    \caption{Average Delivery Time and Energy Consumption with Relative Wind} 
    \label{fig:Energy_wind}
\end{figure*}

\section{Experiments and Results} 

We assess our proposed LOS heuristic composition algorithm compared to flagship algorithms (Dijkstra, Floyd-Warshall, Bellman-Ford), Modified Bellman-Ford (Energy-based Bellman-Ford), and Exhaustive Search. The weight of the edges in flagships is set to be the sum of the travel time on the edge, the recharging time at the next node, and the waiting time caused by preoccupied recharging pads at the next node. The weights of the edges in the Energy-based Bellman-Ford algorithm represent the energy consumed at each edge. We evaluate the algorithms regarding the controlling attributes in a context-aware composition. The controlling attributes include the wind status, carried payload, and recharging pad's availability. The experiments are run on the aforementioned real miniature skyway network. The skyway segments have been scaled up by 50 to mimic the need for recharging stations. We generate 2000 consumer requests with random source and destination and package payloads. We assume that the drone's flight speed is 30 cm/s at the smaller scale and the time it takes to fully charge an empty battery is 40 minutes\footnote{https://www.bitcraze.io/support/f-a-q/}. The drones are assumed to be operating under global wind conditions, i.e., speed and direction. The experiments were carried out on a MacBook Pro, Apple M1 Chip, 8 cores, 16 GB memory, and 1TB SDD.

\textbf{Average Delivery Time.}
In the first experiment, we evaluate the delivery time of the generated requests. The requests are grouped by the number of nodes between the source nodes and the destination nodes of the requests using Dijkstra's algorithm. The x-axis in {\bf Figures \ref{delivery_time}} and {\bf \ref{energy_consumptuin}} represent the number of intermediate nodes. The proposed LOS heuristic algorithm outperforms all algorithms except the Exhaustive search algorithm, as illustrated in { \bf Figure \ref{delivery_time}}. Our proposed algorithm performs well because of its ability to skip nodes when no recharging is required, resulting in shorter delivery times. On the contrary, the other flagship algorithms select paths that may reduce the distance traveled; however, visit nodes with longer transit times. All flagship algorithms compute the same path and delivery times as they behave similarly except when negative edges exist, which is not the case in our scenario. For the Energy-based Bellman-Ford algorithm, the delivery time is still higher than the proposed LOS heuristic but lower than the three flagship algorithms at lower nodes only. The energy-based modified algorithm prefers edges that result in lower recharging times. Lower recharging time refers to edges that are shorter and wind-favored. Therefore, when the number of intermediate nodes is less between the source and destination, the significance of waiting times at each node is not high, resulting in lower delivery times. With more nodes, the waiting time at each node factors in increasing the overall delivery time. In the exhaustive search algorithm, the top 100 shortest paths in terms of distance only are generated. The path with the least delivery time considering all the constraints is selected. Since the network is very dense, i.e., highly connected, generating all possible paths between source and destination is not feasible. Therefore, top $k$ paths are generated. We arbitrarily chose a value of 100 because a lower value of $k$ resulted in unsuccessful compositions. An unsuccessful composition is one in which a drone's battery is depleted before reaching the next node under the current wind and payload conditions. As shown in {\bf Figure \ref{delivery_time}}, the exhaustive search outperforms all the algorithms regarding delivery times as it computes all the possible compositions. However, this performance comes with a high execution time cost, making it infeasible for larger network sizes.

\textbf{Relative Wind Effect on Energy Consumption.} We analyze the effect of the relative wind on energy consumption. We first group the requests by distance, considering the first two groups only, i.e., requests with a total distance traveled between 0-15 km and 15-30 km. Then, within each group, we group the requests with paths that have an average relative wind within 45\degree \ steps between -180\degree \ and 180\degree.
As shown in {\bf Figure \ref{wind-energy}}, when the drone travels with the wind direction (0\degree) or against the wind direction (-180\degree \ and 180\degree), it consumes the least amount of energy. Therefore, if most of the segments composed in a path are in the south or north, the energy consumption is the least. The translational lift increases with a headwind due to the relative airflow increase, resulting in less energy consumption \cite{thibbotuwawa2018energy}. Similarly, the energy consumption is at its lowest with a tailwind because the wind helps reduce the drag forces on the drone. When the relative wind is a crosswind, it consumes more energy than in other directions. We conclude that if two possible paths are available between the source and destination, the path where the drone is oriented to travel with or against the wind is best in terms of energy consumption. This results in shorter recharge and overall delivery times. As shown in {\bf Figure \ref{wind-composition}}, path 2 consumes less energy (190\%) than path 1 (273.5\%) assuming both travel the same distance. This difference in energy consumption is because the drone following path 1 travels a larger distance against the global wind direction than path 2. Furthermore, we observe that the flagship methods slightly outperform the LOS heuristic method in terms of energy consumption, as shown in {\bf Figure \ref{energy_consumptuin}}. This performance of flagship methods is due to selecting paths with shorter distances, where all the energy consumption occurs. The modified Energy-based Bellman-Ford slightly consumes less energy than flagship methods, as it favors paths with less energy consumption. The exhaustive search method outperforms all methods because it selects paths that are both short in distance and wind-favored.

\section{Conclusion}

We present a novel context-aware drone delivery framework using the service paradigm. We model the service as a segment in a skyway network served by a drone. Therefore, a skyway path from source to destination is a service composition consisting of multiple segments served by a drone. We conduct experiments using a real dataset to evaluate our proposed algorithm compared to flagship algorithms and an exhaustive search algorithm. We observe that our proposed algorithm delivers the packages faster than other algorithms while maintaining a similar trend of energy consumption. In the future, we plan to investigate the impact of other environmental factors such as temperature and precipitation on a drone's energy consumption and delivery time.

\section{Acknowledgment}

This research was partly made possible by the LE220100078 and DP220101823 grants from the Australian Research Council. The statements made herein are solely the responsibility of the authors.

\bibliographystyle{IEEEtran}
\bibliography{references}

\begin{thebibliography}{10}
\providecommand{\url}[1]{#1}
\csname url@samestyle\endcsname
\providecommand{\newblock}{\relax}
\providecommand{\bibinfo}[2]{#2}
\providecommand{\BIBentrySTDinterwordspacing}{\spaceskip=0pt\relax}
\providecommand{\BIBentryALTinterwordstretchfactor}{4}
\providecommand{\BIBentryALTinterwordspacing}{\spaceskip=\fontdimen2\font plus
\BIBentryALTinterwordstretchfactor\fontdimen3\font minus
  \fontdimen4\font\relax}
\providecommand{\BIBforeignlanguage}[2]{{%
\expandafter\ifx\csname l@#1\endcsname\relax
\typeout{** WARNING: IEEEtran.bst: No hyphenation pattern has been}%
\typeout{** loaded for the language `#1'. Using the pattern for}%
\typeout{** the default language instead.}%
\else
\language=\csname l@#1\endcsname
\fi
#2}}
\providecommand{\BIBdecl}{\relax}
\BIBdecl

\bibitem{7807172}
F.~F. Mueller and A.~Schmidt, ``Drones ripe for pervasive use,'' \emph{IEEE
  Pervasive Computing}, pp. 21--23, 2017.

\bibitem{9707164}
B.~Alkouz, B.~Shahzaad, and A.~Bouguettaya, ``Service-based drone delivery,''
  in \emph{IEEE CIC}, 2021, pp. 68--76.

\bibitem{chmaj2015distributed}
G.~Chmaj and H.~Selvaraj, ``Distributed processing applications for
  {UAV}/drones: A survey,'' in \emph{Progress in Systems Engineering}, 2015,
  pp. 449--454.

\bibitem{KUMAR20211}
A.~Kumar \emph{et~al.}, ``A drone-based networked system and methods for
  combating coronavirus disease ({COVID-19}) pandemic,'' \emph{Future
  Generation Computer Systems}, pp. 1--19, 2021.

\bibitem{Aurambout2019}
J.-P. Aurambout, K.~Gkoumas, and B.~Ciuffo, ``Last mile delivery by drones: an
  estimation of viable market potential and access to citizens across european
  cities,'' \emph{European Transport Research Review}, pp. 1--21, 2019.

\bibitem{9380171}
A.~Hamdi \emph{et~al.}, ``Drone-as-a-service composition under uncertainty,''
  \emph{IEEE Transactions on Services Computing}, pp. 2685--2698, 2022.

\bibitem{shahzaad2021top}
B.~Shahzaad and A.~Bouguettaya, ``Top-k dynamic service composition in skyway
  networks,'' in \emph{ICSOC}, 2021, pp. 479--495.

\bibitem{10.1145/3460418.3479289}
W.~Lee \emph{et~al.}, ``Package delivery using autonomous drones in skyways,''
  in \emph{Proc. UbiComp/ISWC}, 2021, p. 48–50.

\bibitem{5}
R.~D'Andrea, ``Guest editorial can drones deliver?'' \emph{IEEE Transactions on
  Automation Science and Engineering}, pp. 647--648, 2014.

\bibitem{cokyasar2021designing}
T.~Cokyasar, W.~Dong, M.~Jin, and {\.I}.~{\"O}. Verbas, ``Designing a drone
  delivery network with automated battery swapping machines,'' \emph{Computers
  \& Operations Research}, vol. 129, p. 105177, 2021.

\bibitem{galkin2019uavs}
B.~Galkin, J.~Kibilda, and L.~A. DaSilva, ``{UAVs} as mobile infrastructure:
  Addressing battery lifetime,'' \emph{IEEE Communications Magazine}, pp.
  132--137, 2019.

\bibitem{huang2021deployment}
H.~Huang and A.~V. Savkin, ``Deployment of charging stations for drone delivery
  assisted by public transportation vehicles,'' \emph{IEEE Transactions on
  Intelligent Transportation Systems}, vol.~23, no.~9, pp. 15\,043--15\,054,
  2021.

\bibitem{jermaine2021demo}
J.~Janszen \emph{et~al.}, ``Constraint-aware trajectory for drone delivery
  services,'' in \emph{ICSOC}, 2022, pp. 306--310.

\bibitem{8798337}
G.~Brunner \emph{et~al.}, ``The urban last mile problem: Autonomous drone
  delivery to your balcony,'' in \emph{ICUAS}, 2019, pp. 1005--1012.

\bibitem{8818436}
B.~{Shahzaad} \emph{et~al.}, ``Composing drone-as-a-service ({DaaS}) for
  delivery,'' in \emph{IEEE ICWS}, 2019, pp. 28--32.

\bibitem{grippa2019drone}
P.~Grippa \emph{et~al.}, ``Drone delivery systems: Job assignment and
  dimensioning,'' \emph{Autonomous Robots}, pp. 261--274, 2019.

\bibitem{chu2021simulation}
T.~Chu \emph{et~al.}, ``Simulation and characterization of wind impacts on suas
  flight performance for crash scene reconstruction,'' \emph{Drones}, pp.
  1--23, 2021.

\bibitem{kang2020computational}
G.~Kang, J.-J. Kim, and W.~Choi, ``Computational fluid dynamics simulation of
  tree effects on pedestrian wind comfort in an urban area,'' \emph{Sustainable
  Cities and Society}, pp. 1--17, 2020.

\bibitem{thibbotuwawa2018energy}
A.~Thibbotuwawa \emph{et~al.}, ``Energy consumption in unmanned aerial
  vehicles: A review of energy consumption models and their relation to the
  {UAV} routing,'' in \emph{ISAT}, 2018, pp. 173--184.

\end{thebibliography}

\begin{IEEEbiography}{Babar Shahzaad}{\,}is a Research Fellow in the School of Information Systems at Queensland University of Technology (QUT). He has completed his Ph.D. in Computer Science from The University of Sydney. He has published in top-ranked conferences and journals, including IEEE ICWS, ICSOC, IEEE IoT, and FGCS. His research interests include the Industrial Internet of Things, ICN/NDN applications for the Internet of Things (IoT), Service Computing, and Drone-based Delivery Services in Smart Cities.
Contact him at \href{mailto:babar.shahzaad@sydney.edu.au}{babar.shahzaad@sydney.edu.au}
\end{IEEEbiography}

\begin{IEEEbiography}
{Balsam Alkouz}{\,}is a Ph.D. student in the School of Computer Science at the University of Sydney. She received her bachelor's degree in IT Multimedia and her master’s degree in Computer Science from the University of Sharjah, United Arab Emirates, in 2016 and 2018 respectively. She worked as a Research Assistant in the Data Mining and Multimedia Research Group at the University of Sharjah. Her research focuses on IoT, Service Computing, and Data Mining. Contact her at \href{mailto:balsam.alkouz@sydney.edu.au}{balsam.alkouz@sydney.edu.au}

\end{IEEEbiography}

\begin{IEEEbiography}{Jermaine Janszen}{\,}is a Software Engineer at WiseTech Global. He completed his Honours degree under the supervision of Prof. Athman Bouguettaya at the University of Sydney. His interests focus on Drone-based Services in smart cities.
Contact him at \href{mailto:jjan3640@uni.sydney.edu.au}{jjan3640@uni.sydney.edu.au}
\end{IEEEbiography}

\begin{IEEEbiography}
{Athman Bouguettaya}{\,}is a Professor in the School of Computer Science at the University of Sydney. He received his Ph.D. in Computer Science from the University of Colorado at Boulder (USA) in 1992. He is or has been on the editorial boards of several journals, including the IEEE Transactions on Services Computing, ACM Transactions on Internet Technology, the International Journal on Next Generation Computing, and the VLDB Journal. He is a Fellow of the IEEE and a Distinguished Scientist of the ACM. He is a member of the Academia Europaea (MAE).
Contact him at \href{mailto:athman.bouguettaya@sydney.edu.au}{athman.bouguettaya@sydney.edu.au}
\end{IEEEbiography}

\end{document}